\documentclass{article} 
\usepackage{iclr2023_conference_modified,times}

\usepackage{amsmath,amsfonts,bm}

\def\eqref#1{equation~\ref{#1}}

\def\1{\bm{1}}

\DeclareMathAlphabet{\mathsfit}{\encodingdefault}{\sfdefault}{m}{sl}
\SetMathAlphabet{\mathsfit}{bold}{\encodingdefault}{\sfdefault}{bx}{n}

\usepackage{hyperref}
\usepackage{url}

\usepackage{tikz}
\usepackage{pgfplots}
\usetikzlibrary{shapes.arrows, patterns, calc}
\usepackage{tikz-3dplot}
\usepackage{bbm}
\ifpdf
  \DeclareGraphicsExtensions{.eps,.pdf,.png,.jpg}
\else
  \DeclareGraphicsExtensions{.eps}
\fi

\definecolor{lightblue}{HTML}{A1BDC7}
\definecolor{orange}{HTML}{D98C21}
\definecolor{silver}{HTML}{B0ABA8}
\definecolor{rust}{HTML}{B8420F}
\definecolor{seagreen}{HTML}{2E6B69}
\definecolor{joshua}{HTML}{FBDC7F}
\definecolor{darksky}{HTML}{154c79}

\colorlet{lightsilver}{silver!30!white}
\colorlet{darkorange}{orange!85!black}
\colorlet{darksilver}{silver!85!black}
\colorlet{darklightblue}{lightblue!85!black}
\colorlet{darkrust}{rust!85!black}
\colorlet{darkseagreen}{seagreen!85!black}

\usepackage{graphicx}

\usepackage{hyperref}
\usepackage{url}
\usepackage{float}
\usepackage{array,multirow,graphicx}
\usepackage{float}
\usepackage{mathtools}
\usepackage{amssymb}
\usepackage{amsmath}
\usepackage{caption} 
\usepackage{comment}

\usepackage{xcolor}
\definecolor{lightblue}{HTML}{A1BDC7}
\definecolor{seagreen}{HTML}{2E6B69}
\definecolor{orange}{HTML}{D98C21}
\definecolor{silver}{HTML}{B0ABA8}
\definecolor{rust}{HTML}{B8420F}
\definecolor{joshua}{HTML}{FBDC7F}
\definecolor{darksky}{HTML}{154c79}

\colorlet{lightsilver}{silver!30!white}
\colorlet{darkorange}{orange!85!black}
\colorlet{darksilver}{silver!85!black}
\colorlet{darklightblue}{lightblue!85!black}
\colorlet{darkrust}{rust!85!black}
\colorlet{darkseagreen}{seagreen!85!black}

\newcommand{\Reals}{\mathbb{R}}

\newcommand{\boundx}{\overline{x}}
\newcommand{\boundy}{\overline{y}}
\newcommand{\bboundx}{\boldsymbol{\overline{x}}}

\newcommand{\cA}{\mathcal{A}}
\newcommand{\cP}{\mathcal{P}}
\newcommand{\cD}{\mathcal{D}}

\newcommand{\bd}{\boldsymbol{\delta}}
\newcommand{\bp}{\boldsymbol{\pi}}

\newcommand{\nd}{\delta}
\newcommand{\np}{\pi}

\newcommand{\bx}{\boldsymbol{x}}

\newcommand{\bA}{\boldsymbol{A}}
\newcommand{\bff}{\boldsymbol{f}}
\newcommand{\bzero}{\boldsymbol{0}}

\newcommand{\PINNLoss}{\mathcal{L}^{\text{PINN}}}
\newcommand{\PINNLossO}{\mathcal{L}_{0}^{\text{PINN}}}
\newcommand{\PINNBCLoss}{\mathcal{L}_{\partial \Omega}^{\text{PINN}}}

\newcommand{\PINNPDELoss}{\mathcal{L}_{ \Omega}^{\text{PINN}}}

\newcommand{\CPINNLoss}{\mathcal{L}^{\text{CPINN}}}
\newcommand{\CPINNLossO}{\mathcal{L}_{0}^{\text{CPINN}}}
\newcommand{\CPINNBCLoss}{\mathcal{L}_{\partial\Omega}^{\text{CPINN}}}
\newcommand{\CPINNBCLossPrime}{\mathcal{L}_{\partial\Omega^\prime}^{\text{CPINN}}}

\newcommand{\CPINNPDELoss}{\mathcal{L}_{\Omega}^{\text{CPINN}}}
\DeclareMathOperator{\sech}{sech}

\hypersetup{colorlinks=true,linkcolor=darkrust,citecolor=darklightblue,urlcolor=darksilver}

\usepackage[utf8]{inputenc} 
\usepackage[T1]{fontenc}    
\usepackage{hyperref}       
\usepackage{url}            
\usepackage{booktabs}       
\usepackage{amsfonts}       
\usepackage{nicefrac}       
\usepackage{microtype}      
\usepackage{cleveref}

\begin{document}

%

\title{Competitive Physics Informed Networks}

%

\author{Qi Zeng, Yash Kothari, Spencer H. Bryngelson \& Florian Sch{\"a}fer \\
School of Computational Science and Engineering\\
Georgia Institute of Technology\\
Atlanta, GA 30332, USA \\
\texttt{\{qzeng37@,ykothari3@, shb@,florian.schaefer@cc.\}gatech.edu} \\
}
%

\newcommand{\fix}{\marginpar{FIX}}
\newcommand{\new}{\marginpar{NEW}}

\iclrfinalcopy 

\maketitle

\begin{abstract}
    Neural networks can be trained to solve partial differential equations (PDEs) by using the PDE residual as the loss function.
    This strategy is called ``physics-informed neural networks'' (PINNs), but it currently cannot produce high-accuracy solutions, typically attaining about $0.1\%$ relative error.
    We present an adversarial approach that overcomes this limitation, which we call competitive PINNs (CPINNs).
    CPINNs train a discriminator that is rewarded for predicting mistakes the PINN makes.
    The discriminator and PINN participate in a zero-sum game with the exact PDE solution as an optimal strategy.
    This approach avoids squaring the large condition numbers of PDE discretizations, which is the likely reason for failures of previous attempts to decrease PINN errors even on benign problems.
    Numerical experiments on a Poisson problem show that CPINNs achieve errors four orders of magnitude smaller than the best-performing PINN.
    We observe relative errors on the order of single-precision accuracy, consistently decreasing with each epoch. 
    To the authors' knowledge, this is the first time this level of accuracy and convergence behavior has been achieved. 
    Additional experiments on the nonlinear Schr{\"o}dinger, Burgers', and Allen--Cahn equation show that the benefits of CPINNs are not limited to linear problems.
\end{abstract}

\section{Introduction}\label{sec:intro}

\textbf{PDE-constrained deep learning.} 
Partial differential equations (PDEs) model physical phenomena like fluid dynamics, heat transfer, electromagnetism, and more.
The rising interest in scientific machine learning motivates the study of PDE-constrained neural network training \citep{lavin2021simulation}.
Such methods can exploit physical structure for learning or serve as PDE solvers in their own right.

\textbf{Physics informed networks.} 
\citet{lagaris1998artificial} represent PDE solutions as neural networks by including the square of the PDE residual in the loss function, resulting in a neural network-based PDE solver.
\citet{RAISSI2019686} recently refined this approach further and called it ``physics informed neural networks (PINNs),'' initiating a flurry of follow-up work. 
PINNs are \textit{far} less efficient than classical methods for solving most PDEs, but are promising tools for high-dimensional or parametric PDEs \citep{xue2020amortized} and data assimilation problems.
The training of PINNs also serves as a model problem for the general challenge of imposing physical constraints on neural networks, an area of fervent and increasing interest \citep{wang2021learning,li2021physics,donti2021dc}. 

\textbf{Training pathologies in PINNs.}
PINNs can, in principle, be applied to all PDEs, but their numerous failure modes are well-documented~\citep{wang2020understanding,liu2021physicsaugmented,krishnapriyan2021characterizing}.
For example, they are often unable to achieve high-accuracy solutions.
The first works on PINNs reported relative $L_2$ errors of about $10^{-3}$~\citep{RAISSI2019686}. 
The authors are unaware of PINNs achieving errors below $10^{-5}$, even in carefully crafted, favorable settings.
Higher accuracy is required in many applications.

\textbf{Existing remedies.} 
A vast and growing body of work aims to improve the training of PINNs, often using problem-specific insights. 
For example, curriculum learning can exploit causality in time-dependent PDEs \citep{krishnapriyan2021characterizing,wang2022respecting,wight2020solving}.
\citet{krishnapriyan2021characterizing} also design curricula by embedding the PDE in a parametric family of problems of varying difficulty. 
Other works propose adaptive methods for selecting the PINN collocation points \citep{lu2021deepxde,nabian2021efficient,daw2022rethinking}.
Adaptive algorithms for weighing components of the PINN loss function have also been proposed \citep{wang2022and,van2022optimally}. 
Despite these improvements, the squared residual penalty method used by such PINNs imposes a fundamental limitation associated with conditioning, which is discussed next.

\textbf{The key problem: Squared residuals.} 
Virtually all PINN-variants use the squared PDE residual as loss functions.
For a linear PDE of order $s$, this is no different than solving an equation of order $2s$, akin to using normal equations in linear algebra.
The condition number $\kappa$ of the resulting problem is thus the \emph{square} of the condition number of the original one.
Solving discretized PDEs is an ill-conditioned problem, inhibiting the convergence of iterative solvers and \textit{explaining the low accuracy of most previous PINNs}.
It is tempting to address this problem using penalties derived from $p$-norms with $p \neq 2$.
However, choosing $p > 2$ leads to worse condition numbers, whereas $p < 2$ sacrifices the smoothness of the objective. 
The convergence rates of gradient descent on (non-)smooth convex problems suggest that this trade is unfavorable \citep{bubeck2015convex}.

\textbf{Weak formulations.} 
Integration by parts allows the derivation of a weak form of a PDE, which for some PDEs can be turned into a minimization formulation that does not square the condition number.
This procedure has been successfully applied by~\citet{yu2017deep} to solve PDEs with neural networks (\emph{Deep Ritz}). 
However, the derivation of such minimization principles is problem-dependent, limiting the generality of the formulation.
Deep Ritz also employs penalty methods to enforce boundary values, though these preclude the minimization problem's solution from being the PDE's exact solution.
\citet{liao2019deep} proposed a partial solution to this problem.
The work most closely related to ours is by \citet{zang2020weak}, who proposes a game formulation based on the weak form.

\textbf{Competitive PINNs.}
We propose \textit{Competitive} Physics Informed Neural Networks (CPINNs) to address the above problems.
CPINNs are trained using a minimax game between the PINN and a discriminator network.
The discriminator learns to predict mistakes of the PINN and is rewarded for correct predictions, whereas the PINN is penalized.
We train both players simultaneously on the resulting zero-sum game to reach a Nash equilibrium that matches the exact solution of the PDE.

\textbf{Summary of contributions.}
A novel variant of PINNs, called CPINNs, is introduced, replacing the penalty employed by PINNs with a primal-dual approach.
This simultaneously optimizes the PINN and a discriminator network that learns to identify violations of the PDE and boundary constraints.
We optimize PINNs with competitive gradient (CGD)~\citep{schafer2020competitive} and compare their performance to regular PINNs trained with Adam. 
On a two-dimensional Poisson problem, CPINNs achieve a relative accuracy of almost $10^{-8}$, improving over PINNs by four orders of magnitude. 
To the best of our knowledge, this is the first time a PINN-like network was trained to this level of accuracy. 
We compare PINNs with CPINNs on a nonlinear Schr{\"o}dinger equation, a viscous Burgers' equation, and an Allen-Cahn equation. 
In all but the last case, CPINNs improve over PINNs trained with a comparable computational budget.

\section{Competitive PINN formulation}

We formulate CPINNs for a PDE of the general form 
\begin{align}
    \mathcal{A}[u] &= f, \quad \text{in}\ \Omega \label{e:pde} \\
    u &= g,\quad \text{on}\ \partial\Omega, \label{e:boundary}
\end{align}
where $\mathcal{A}[\cdot]$ is a (possibly nonlinear) differential operator and  $\Omega$ is a domain in $\Reals^d$ with boundary $\partial \Omega$.
To simplify notation, we assume that $f$, $g$, and $u$ are real-valued functions on $\Omega$, $\partial \Omega$, and $\Omega \cup \partial \Omega$, respectively.
One can extend both PINNs and CPINNs to vector-valued such functions if needed.

\subsection{Physics Informed Neural Networks (PINNs)}

PINNs approximate the PDE solution $u$ by a neural network $\mathcal{P}$ mapping $d$-variate inputs to real numbers.
The weights are chosen such as to satisfy~\eqref{e:pde} and~\eqref{e:boundary} on the points $\bx \subset \Omega$ and $\bboundx \subset \partial \Omega$.
The loss function used to train $\cP$ has the form
\begin{gather}
    \PINNLoss(\cP, \bx, \bboundx) = 
    \PINNPDELoss (\cP, \bx_\Omega) + \PINNBCLoss (\cP, \bboundx),
    \label{e:PINNloss}
\end{gather}
where $\PINNBCLoss$ measures the violation of the boundary conditions~\eqref{e:boundary} and $\PINNPDELoss$ measures the violation of the PDE of~\eqref{e:pde}.
They are defined as
\begin{align}
    \PINNPDELoss (\cP, \bx) 
        &= \frac{1}{N_{\Omega}} \sum_{i=1}^{N_{\Omega}} 
        \left(\mathcal{A}[\cP](x_{i})) - f(x_{i}) \right)^2\\
    \PINNBCLoss (\cP, \bboundx) 
        &= \frac{1}{N_{\partial\Omega}}\sum_{i = 1}^{N_{\partial\Omega} }
        \left( \cP \left( \boundx_i \right) - g \left( \boundx_i \right) \right)^2.
\end{align}
Here, $N_{\Omega}$ and $N_{\partial\Omega}$ are the number of points in the sets $\bx$ (interior) and $\bboundx$ (boundary), and $x_i$ and $\boundx_{i}$ are the $i$-th such points in $\bx$ and $\bboundx$.

PINNs approximate the exact solution $u$ of the PDE by minimizing the loss in \eqref{e:PINNloss}. 
However, optimizing this loss using established methods such as gradient descent, ADAM, or LBFGS often leads to unacceptably large errors or an inability to train at all~\citep{wang2020understanding, wang2022and}.
This pathology has been attributed to the bad conditioning of the training problem \citep{krishnapriyan2021characterizing}.
Next, we introduce CPINNs, a game-based formulation designed to mitigate these problems.

\subsection{Competitive Physics Informed Neural Networks (CPINNs)}

CPINNs introduce one or more discriminator networks $\cD$ with input $x \in \Reals^d$ and outputs $\cD_{\Omega}(x)$ and $\cD_{\partial \Omega}(x)$. 
$\cP$ and $\cD$ compete in a zero-sum game where $\cP$ learns to solve the PDE, and $\cD$ learns to predict the mistakes of $\cP$.
This game is defined as a minimax problem
\begin{gather}
   \max_{\cD}\min_{\cP} \CPINNPDELoss(\cP, \cD, \bx) + \CPINNBCLoss(\cP, \cD, \bboundx),
   \label{e:minimax}
\end{gather}
where
\begin{gather}
    \CPINNPDELoss(\cD,\cP,\bx) = 
    \frac{1}{N_{\Omega}}\sum_{i=1}^{N_\Omega} \cD_{\Omega} (x_{i})
    \left( 
        \cA[\cP](x_{i}) -f(x_{i}) 
    \right),  \\
    \CPINNBCLoss (\cD,\cP, \bboundx) = 
    \frac{1}{N_{\partial \Omega}}\sum_{i=1}^{N_{\partial \Omega}} 
    \cD_{\partial \Omega}(\boundx_{i})
    \left(
        \cP\left( \boundx_{i}\right) - g \left(\boundx_i \right)  
    \right).
\end{gather} 
Here, $\cD_{\Omega}(x_i)$ and $\cD_{\partial \Omega}(\boundx)$ can be interpreted as \textit{bets} by the discriminator that the PINN will over- or under-shoot~\eqref{e:pde} and~\eqref{e:boundary}.
Winning the bet results in a reward for $\cD$ and a penalty for $\cP$, and a lost bet has the opposite effect. 

The Nash equilibrium of this game is $\cP \equiv u$ and $\cD \equiv 0$.
Thus, iterative algorithms for computing Nash equilibria in such zero-sum games can be used to solve the PDE approximately.
This work focuses on the CGD algorithm of \citet{schafer2020competitive}.
Still, CPINNs can be trained with other proposed methods for smooth game optimization~\citep{korpelevich1977extragradient,mescheder2017numerics,balduzzi2018mechanics,gemp2018global,daskalakis2018training,letcher2019differentiable}.
In our experiments, $\cP$ and $\cD$ are fully connected networks with hyperbolic tangent and ReLU activation functions, respectively. 
Each network's number of layers and neurons depends on the PDE problem.

\subsection{Avoiding squares of differential operators}\label{sec:cpinn_advantage}

Multiagent methods for solving PDEs may seem unorthodox, yet they are motivated by observations in classical numerical analysis.
Consider the particular case of a linear PDE and networks $\cP, \cD$ with outputs that depend linearly on their weight-vectors $\bp$ and $\bd$ resulting in the parametric form  
\begin{equation}
    \cP(x) = \sum_{i = 1}^{\dim(\bp)} \np_i \psi_i(x), \quad   \cD(x) = \sum_{i = 1}^{\dim(\bd)} \nd_i \phi_i(x),
\end{equation}
for basis function sets $\left\{\psi_i\right\}_{1 \leq i \leq \dim(\bp)}$ and $\left\{\phi_i\right\}_{1 \leq i \leq \dim(\bd)}$.
We focus our attention on the PDE constraint in~\eqref{e:pde}, which is evaluated at a set $\bx$ of $N_{\Omega}$ points.
Defining $\bA \in \Reals^{N_{\Omega} \times \dim(\bp)}$ and $\bff \in \Reals^{N_{\Omega}}$, this leads to
\begin{equation}
    A_{ij} \coloneqq \cA[\psi_j](x_i), \quad f_i \coloneqq f(x_i)
\end{equation}
and the discretized PDE 
\begin{equation}
    \bA \bp = \bff.
    \label{e:discretePDE}
\end{equation}
PINNs solve~\eqref{e:discretePDE} via a least squares problem 
\begin{gather}
    \min \limits_{\bp} \| \bA \bp - \bff \|^2,
    \label{e:leastsquares}
\end{gather}
trading the equality constraint for a minimization problem with solution $\bp = (\bA^{\top} \bA)^{-1} \bA^{\top} \bff$.

Since the matrix $(\bA^{\top} \bA)$ is symmetric positive-definite, one can solve it with specialized algorithms such as the conjugate gradient method (CG)~\citep{shewchuk1994introduction}. 
This approach is beneficial for well-conditioned nonsymmetric matrices but inappropriate for ill-conditioned $\bA$~\citep{axelsson1977solution}.
This is because $\kappa(\bA^{\top} \bA) = \kappa(\bA)^2$, resulting in slow convergence of iterative solvers. 
Differential operators are unbounded. 
Thus, their discretization leads to ill-conditioned linear systems. 
\citet{krishnapriyan2021characterizing} argue that the ill-conditioning of~\eqref{e:leastsquares} causes the optimization difficulties they observe.

CPINNs turn the discretized PDE in~\eqref{e:discretePDE} into the saddlepoint problem 
\begin{gather}
    \min_{\bp} \max_{\bd} \bd^\top(\bA \bp - \bff).
    \label{e:matrixminimax}
\end{gather}
The solution to this problem is the same as that of the system of equations
\begin{gather}
\begin{bmatrix}
    \bzero & \bA^\top \\
    \bA & \bzero
\end{bmatrix}
\begin{bmatrix}
\bp \\ \bd
\end{bmatrix} =
\begin{bmatrix}
\bzero \\ \bff
\end{bmatrix}, \quad \text{ with } \quad \kappa\left(\begin{bmatrix}
    \bzero & \bA^\top \\
    \bA & \bzero
\end{bmatrix}\right) = \kappa(\bA).
\label{e:system}
\end{gather}
By turning~\eqref{e:discretePDE} into the saddle point problem of~\eqref{e:matrixminimax} instead of the minimization form of~\eqref{e:leastsquares}, CPINNs avoid squaring the condition number.

In light of the above, deciding between PINNs and CPINNs is the nonlinear analog of an old dilemma in numerical linear algebra: should one trade solving a linear system for a least-squares problem?  
This trade allows using simpler algorithms like CG at the cost of a squared condition number.
Yet, it is rarely beneficial for the ill-conditioned systems arising from discretized PDEs.
As demonstrated in the following experiments, the complexity of solving multiagent problems is a price worth paying.

\section{Results}\label{sec:results}

\subsection{Overview} 
This section compares PINNs and CPINNs on a series of model problems. 
The code used to produce the experiments described below can be found under \url{github.com/comp-physics/CPINN}.
We study a two-dimensional Poisson problem (\cref{s:poisson}), a nonlinear Schr{\"o}dinger equation (\cref{s:SchrodingerSection}), a viscous Burgers' equation (\cref{s:BurgersSection}), and the Allen-Cahn equation (\cref{s:ACsection}). 
Throughout the previous sections, we train PINNs using Adam and CPINN using adaptive competitive gradient descent (ACGD) \citep{schafer2020implicit}. 
The latter combines CGD of \citet{schafer2020competitive} with an Adam-like heuristic for choosing step sizes.

ACGD uses GMRES \citep{saad1986gmres} and the Hessian vector products obtained by automatic differentiation to solve the linear system defining the CGD update.
Thus, iterations of (A)CGD are considerably more expensive than those of Adam. 
To account for this difference fairly, we also provide the number of forward passes through the neural network required by the two methods.
An Adam iteration amounts to a single forward pass and an ACGD iteration to two forward passes, plus two times the number of GMRES iterations.
We also tried other optimizers for both PINNs and CPINNs but found them inferior to Adam and ACGD.
A comparison is presented in \cref{s:optimizers}.

\begin{figure}
    \centering
    \includegraphics[scale=1]{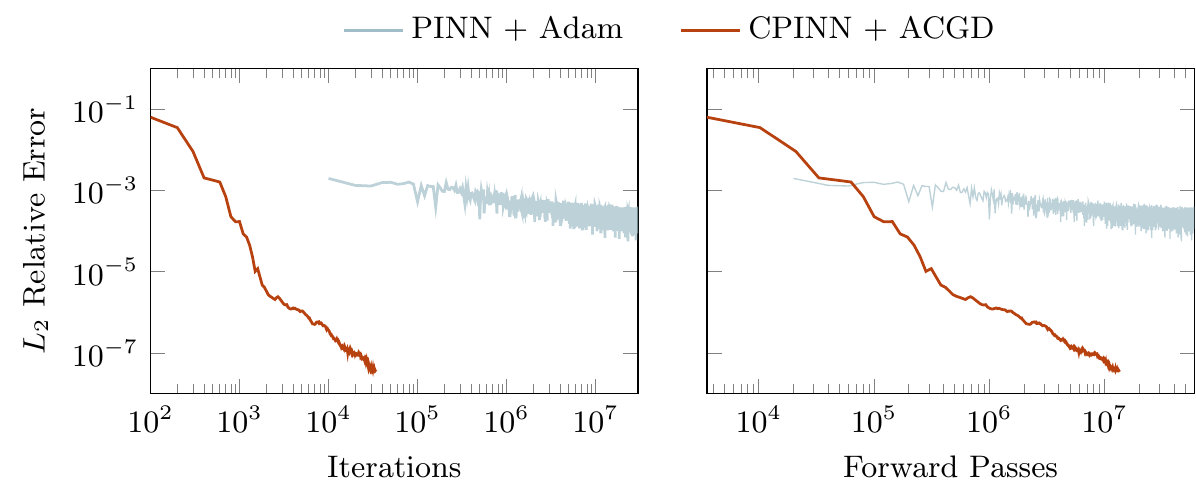}
    \caption{
        Comparison of CPINN and PINN on the Poisson problem of \eqref{e:poisson} in terms of relative error.
        CPINN has a faster convergence rate and reduces the $L_2$ error to $1.7\times 10^{-8}$, whereas the PINN case has an $L_2$ error of $1.2 \times 10^{-4}$ even with a larger computational budget. 
    } 
    \label{fig:poisson}
\end{figure}

\subsection{Poisson Equation}\label{s:poisson}

We begin by considering a two-dimensional Poisson equation:
\begin{gather}
    \Delta u(x, y) = -2 \sin(x) \cos(y), \quad x,y \in [-2,2] 
    \label{e:poisson}
\end{gather}
with Dirichlet boundary conditions
\begin{align*}
    u(x,-2)            &= \sin(x)\cos(-2),            \qquad  u(-2,y) = \sin(-2)\cos(y), \\
    u(x,\phantom{-}2)  &= \sin(x)\cos(\phantom{-}2),  \qquad  u(\phantom{-}2,y)  = \sin(\phantom{-}2)\cos(y).
\end{align*}
This problem has the manufactured solution
\begin{gather}
    u(x,y) = \sin(x) \cos(y).
\end{gather}
The PINN implementation has 
losses
\begin{align}
    \PINNBCLoss  &= 
        \frac{1}{N_{\partial \Omega}} \sum_{i=1}^{N_{\partial \Omega}}
        \left( \cP(\boundx_i, \boundy_i) - u(\boundx_i, \boundy_i) \right)^2 , \\
    \PINNPDELoss &= 
        \frac{1}{N_{\Omega}} \sum_{i=1}^{N_{\Omega}} 
        \left( \cP_{xx} (x_{i}, y_{i}) + \cP_{yy}(x_{i}, y_{i}) + 2 \sin(x_{i}) \cos(y_{i}) 
        \right)^2,
\end{align}
and the CPINN losses are 
\begin{align}
    \CPINNBCLoss  &= 
        \frac{1}{N_{\partial \Omega}} \sum_{i=1}^{N_{\partial \Omega}}
         \cD_{\partial \Omega}(\boundx_{i}) \left( \cP(\boundx_i, \boundy_i) - u(\boundx_i, \boundy_i) \right) , \\
    \CPINNPDELoss &= 
        \frac{1}{N_{\Omega}} \sum_{i=1}^{N_{\Omega}} 
        \cD_{\Omega} (x_{i}) \left( \cP_{xx} (x_{i}, y_{i}) + \cP_{yy}(x_{i}, y_{i}) + 2 \sin(x_{i}) \cos(y_{i}) 
        \right).
\end{align}

\begin{figure}
    \centering
    \includegraphics[width=\textwidth, trim=3cm 0 2cm 0]{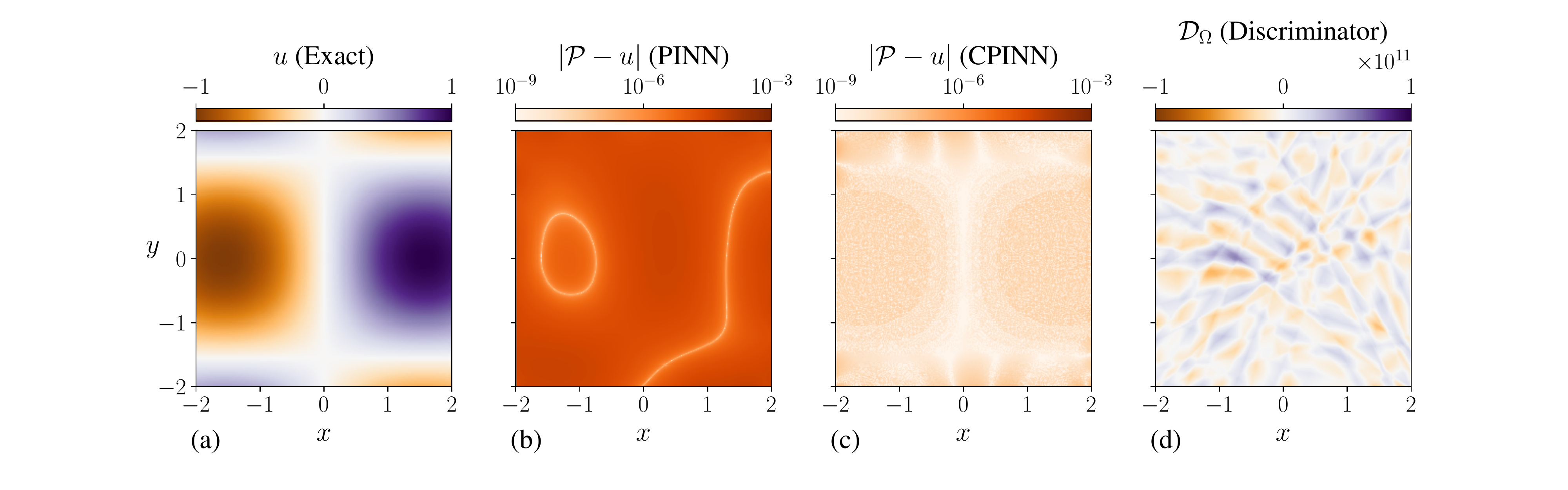}
    \caption{
        (a) Exact solution $u$ to~\eqref{e:pde}, absolute errors of (b) PINN + Adam after $3\times 10^7$ training iterations and (c) CPINN + ACGD after $48\,000$ training iterations, and (d) the discriminator. 
    }
    \label{fig:heatmap}
\end{figure}

\Cref{fig:heatmap}~(a) shows the exact solution of the PDE and the absolute error of the best models trained using (b) PINN and (c) CPINN, as well as an example of the discriminator output (d). 
The CPINN achieves lower errors than the PINN by a factor of about $10^6$ throughout most of the domain. 

\Cref{fig:poisson}~ shows the relative $L_2$ error of a PINN and a CPINN on the Poisson problem. 
CPINN shows a faster convergence rate regarding the number of forward passes and training epochs, and its accuracy is higher than that of PINN by 4 orders of magnitude.

\subsection{Nonlinear Schr\"odinger equation}\label{s:SchrodingerSection}

\begin{figure}
    \centering
    \includegraphics[scale=1]{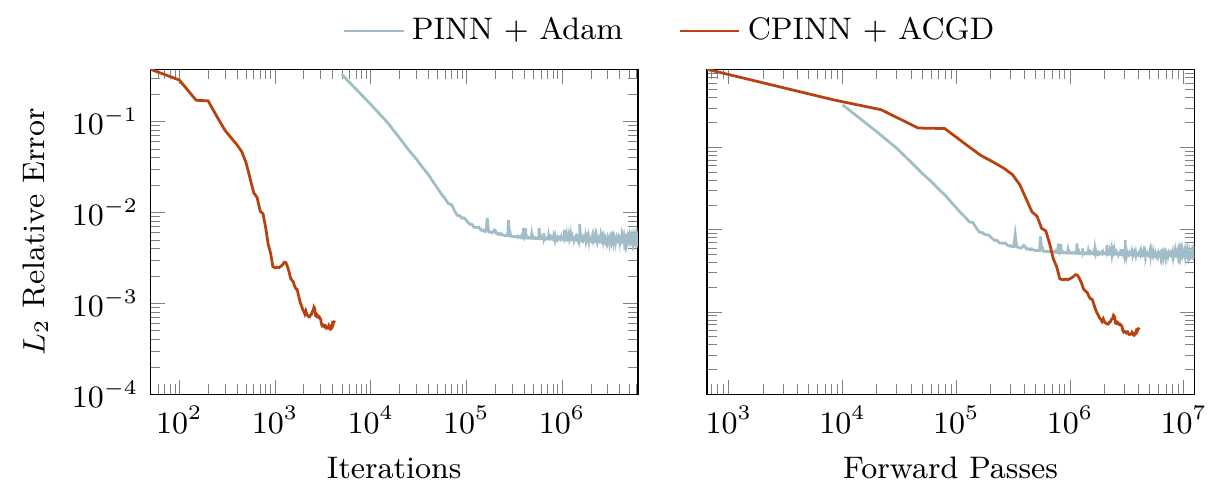}
    \caption{Comparison of CPINN and PINN on the nonlinear Schr\"odinger~\eqref{SchrodingerPDE} in terms of relative errors. 
    After $200\,000$ training iterations, PINN cannot reduce the $L_2$ error further, plateauing about $4 \times 10^{-3}$ , whereas CPINN reduces the error to $6\times 10^{-4}$ under a smaller computational budget.}
    \label{fig:SchrodingerPlot}
\end{figure}

In this subsection, we apply the competitive PINN methodology to solve the Schr\"odinger equation
\begin{gather}
    u_t + \frac{1}{2} u_{xx} + |u|^2 u = 0, \quad x\in [-5, 5], \quad t\in [0, \pi/2], 
    \label{SchrodingerPDE} 
\end{gather}
where $u(t,x)$ is the complex-valued solution and
\begin{gather}
    u(0,x) = 2 \sech(x), \quad u(t, -5) = u(t, 5), \quad u_x(t, -5) = u_x(t, 5)
    \label{SchrodingerBC}
\end{gather}
are the initial and boundary conditions.

The best results for CPINNs and PINNs are presented in \cref{fig:SchrodingerPlot}. 
CPINN reached an $L_2$ relative error of $6\times 10^{-4}$ after $4.2\times 10^3$ training iterations with $4.1\times 10^6$ forward passes.
PINN reached  an $L_2$ relative error of $5\times 10^{-3}$ after $6.2 \times 10^6$ training iterations, equivalent to $6.2 \times 10^6$ forward passes.

\subsection{Burgers' equation}\label{s:BurgersSection}

\begin{figure}
    \centering
    \includegraphics[scale=1]{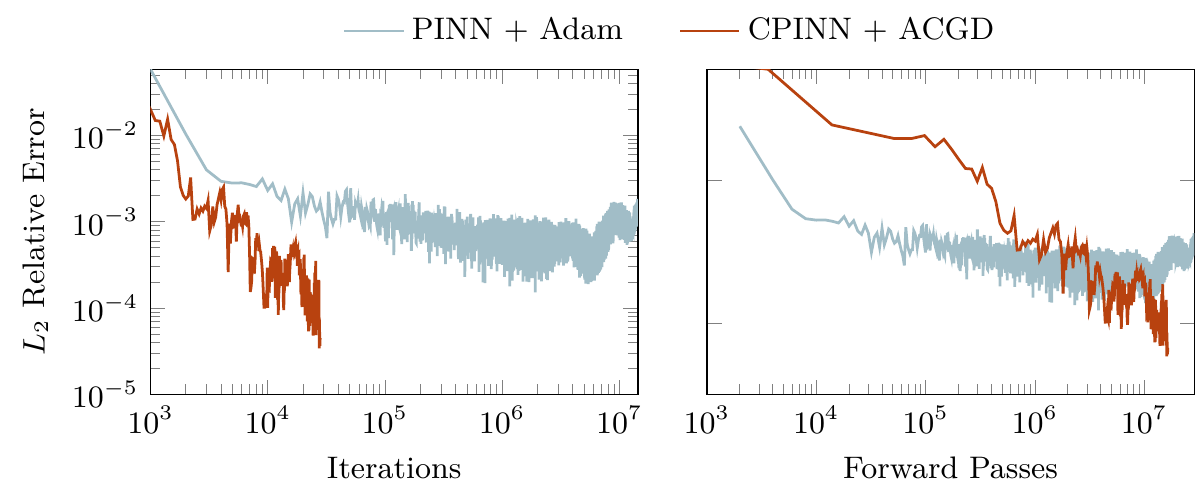}
    \caption{The relative errors of CPINNs (current) and PINNs on the Burgers' equation.}
    \label{fig:burgers}
\end{figure}

We next consider the case of the viscous Burgers' equation. 
This nonlinear second-order equation is
\begin{gather}
    u_t + u u_{x} -  (0.01 / \pi) u_{xx} = 0, \quad x\in [-1, 1], \quad t\in [0, 0], 
    \label{BurgersPDE} 
\end{gather}
which has parameters matching \citet{RAISSI2019686}. $u(t,x)$ is the solution of the PDE and
\begin{gather}
    u(0,x) = -\sin(\pi x), \quad u(t, -1) = u(t, 1) = 0
    \label{BurgersBC}
\end{gather}
are the initial and boundary conditions. 
A comparison to CPINNs is presented in \cref{fig:burgers}. 
CPINN exhibits an improved convergence rate and continues to reduce the error, whereas the progress of PINN eventually stagnates.

\subsection{Allen--Cahn equation}\label{s:ACsection}

We next consider the one-dimensional Allen--Cahn equation with periodic boundary conditions, a cubically nonlinear equation given by 
\begin{gather}
    u_t - 0.0001u_{xx} +  5 u^3 - 5u = 0, \quad x\in [-1, 1], \quad t\in [0, 1], 
    \label{e:ac} 
\end{gather}
where $u(t,x)$ is the solution of the PDE and
\begin{gather}
    u(0,x) = x^2\cos(\pi x), \quad u(t, -1) = u(t, 1), \quad u_x(t, -1) = u_x(t, 1)
    \label{e:acbc}
\end{gather}
are the initial and boundary conditions. 
This example follows \cite{RAISSI2019686}, and we use their training and testing data.

\citet{RAISSI2019686} does not report the performance of a standard PINN and instead shows the performance of a problem-specific modification that exploits the temporal structure of the problem.
\citet{wight2020solving} observe that this problem is difficult to solve directly for PINNs and propose a curriculum learning approach to remedy this problem.

\begin{figure}
    \centering
    \includegraphics[scale=1]{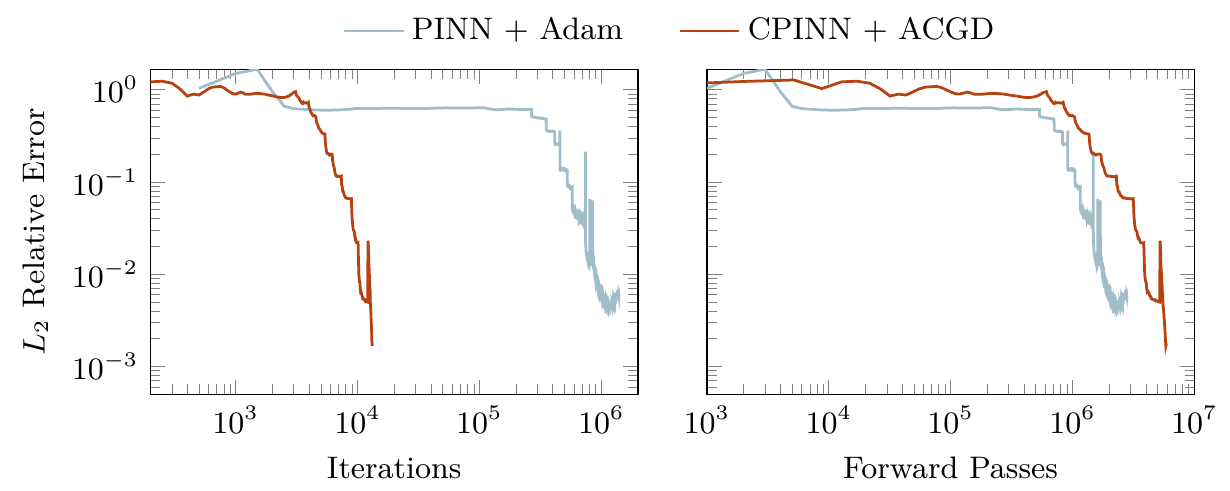}
    \caption{
        Relative error for the Allen--Cahn equation.
        A CPINN and a PINN that uses the \textit{curriculum learning} approach of \cite{wight2020solving} are shown.
        CPINN does not outperform this particular PINN, which may be due to the overall low accuracy of both methods.
    }
    \label{fig:allencahn}
\end{figure}

We use the approach of \citet{wight2020solving} for PINNs and a suitable adaptation for CPINN.
On this problem, CPINN is modestly outperformed by PINNs. 
This may be due to the overall slow convergence that prevents us from reaching the regime where PINNs start to plateau.
It is also possible that our adaption of the curriculum learning approach of \citet{wight2020solving} to CPINNs could be improved.

\begin{figure}
    \centering
    \includegraphics[scale=1]{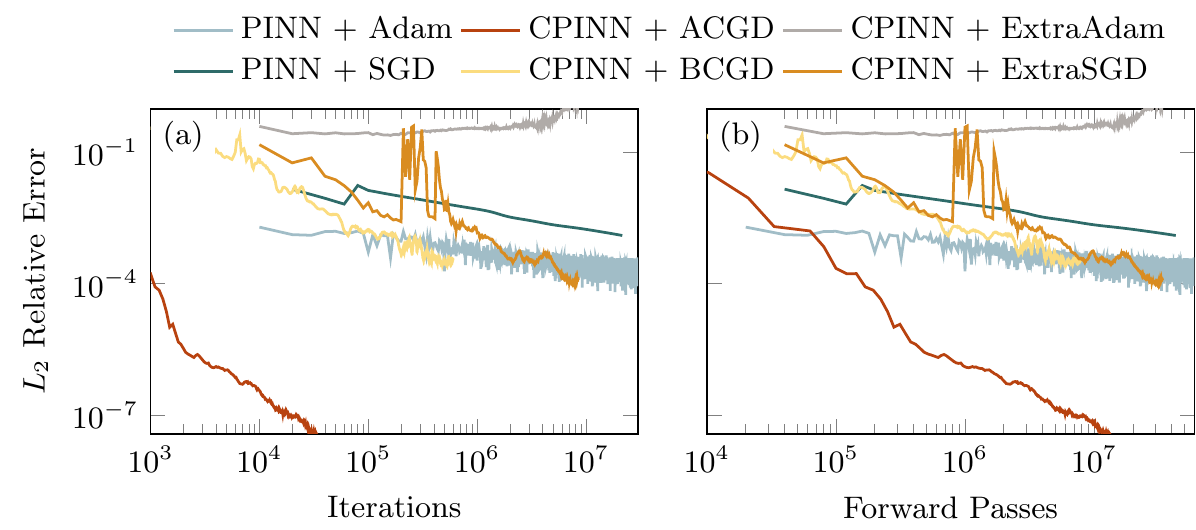}
    \caption{Comparison of PINNs and CPINNs with different optimizers for the Poisson problem.}
    \label{fig:figure}
\end{figure}

\begin{table}
    \centering
    \renewcommand{\arraystretch}{1.2}
    \caption{Performance of CPINNs and PINNs on the 2D Poisson problem of~\eqref{e:poisson}. }\label{tab:CPINNs}
    \begin{tabular}{c c c c c c c} 
         & Optimizer & Iterations & $L_2$ Rel.\ Error & $\PINNLoss$ & $\PINNPDELoss$ &  $\PINNBCLoss$ \\ 
         \hline\hline
         \parbox[t]{2mm}{\multirow{2}{*}{\rotatebox[origin=c]{90}{PINN}}} & Adam & $3 \times 10^7$ & $1.2 \times 10^{-4}$ & $1.4 \times 10^{-8}$ & $8.4\times 10^{-8}$ & $5.5 \times 10^{-9}$ \\
         & SGD &  $2.2 \times 10^7 $& $1.3 \times 10^{-3}$ & $1.1 \times 10^{-6}$ & $5.2 \times 10^{-7}$ & $6.3 \times 10^{-7}$ \\
         \hline
         \parbox[t]{2mm}{\multirow{4}{*}{\rotatebox[origin=c]{90}{CPINN}}} 
         & ACGD & $4.8 \times 10^4$ & $1.7 \times 10^{-8}$ & $2.1 \times 10^{-14}$ &$2 \times 10^{-14}$ & $6 \times 10^{-16}$ \\
         & CGD & $6 \times 10^5$ & $3.5 \times 10^{-4}$ & $5.1\times 10^{-7}$& $3.5 \times 10^{-7}$& $1.7\times 10^{-7}$\\
         & ExtraAdam & $8.5 \times 10^{6}$ & $1.6$ & $1.6\times 10^{6}$ &  $1.6 \times 10^{6}$ & 0.2 \\
         & ExtraSGD & $8.5 \times 10^{6}$ & $1.2 \times 10^{-4}$ & 
         $1.2 \times 10^{-7}$& $9.2 \times 10^{-8}$ & $3 \times 10^{-8}$ \\
         \hline\hline
    \end{tabular}
\end{table}

\subsection{Other optimizers}\label{s:optimizers}

We repeated our experiments with the Poisson problem using different optimizers to distinguish the effects of CPINN and ACGD.
We use Adam and SGD for PINNs, and basic CGD (BCGD, without adaptive step sizes), ACGD, ExtraSGD, and ExtraAdam for CPINNs \citep{korpelevich1977extragradient,gidel2020variational}.
We observe that combined with CPINN, the nonadaptive algorithms BCGD and extragradient improve over PINN trained with (the equally nonadaptive) SGD. 
The latter even begins to improve over PINNs trained with Adam after sufficient iterations. 

We could not achieve training via ExtraAdam on the CPINN strategy.
The ACGD optimizer achieves much better results than all others. 
We conclude that CPINN improves over PINN, but the full realization of the CPINN advantage is enabled by the robust and adaptive training afforded by ACGD.

\section{Connections to existing work}

Saddle point formulations arise in Petrov--Galerkin and mixed finite-element methods, which use different finite-dimensional function spaces to represent the solution (called trial functions) and measure the violation of the equations (called test functions)~\citep{quarteroni2008numerical,fortin1991mixed}.
The neural networks introduced here, $\cP$ and $\cD$, are nonlinear analogs of trial and test spaces. 
Unlike our approach, \textit{Deep Petrov--Galerkin} of \citet{shang2022deep} only parameterizes the trial space by a neural network and uses a conventional discretization such as finite elements for the test space. 
They only train the last layer while keeping all other layers fixed after initialization.

Saddle point problems in PDEs can be interpreted geometrically and have a game-theoretic interpretation~\citep{lemaire1973saddle}. 
In a complementary game-theoretic approach, \citet{owhadi2017multigrid} cast computation as a game between a numerical analyst and an environment, using ideas from decision theory~\citep{wald1945statistical}. 
Saddle point problems also arise from the introduction of Lagrange multipliers, which are used to cast a constrained optimization problem as an unconstrained saddle point problem~\citep{brezzi1974existence}. 
These discriminators can be viewed as neural-network-parametrized Lagrange multipliers that enforce distributional equality (in GANs) or satisfaction of the PDE (in CPINNs).

Our work is thus related to recent efforts that combine CGD with Lagrange multipliers to solve constrained optimization problems arising from reinforcement learning and computer graphics \citep{bacon2019lagrangian,yu2021robust,soliman2021constrained}.
The need for constrained training of neural networks also arises in other applications, resulting in an increasing body of work on this topic~\citep{pathak2015constrained,donti2021dc,lokhande2020fairalm,fioretto2020lagrangian}. 

Following \cref{sec:intro}, \emph{Deep Ritz} exploits the fact that some PDEs can be cast as minimization problems without squaring the condition number~\citep{yu2017deep}, expressing~\eqref{e:discretePDE} as 
\begin{gather*}
    \min_{\bp} \frac{\bp^{\top} \bA \bp}{2} - \bp^{\top} \bff,
\end{gather*} 
for a symmetric positive definite matrix $\bA$.
Such a formulation is not always available, and even if it exists, one still must enforce the PDE boundary conditions~\citep{liao2019deep}.
\citet{zang2020weak} uses a weak PDE formulation to derive a minimax formulation. 
In the notation of \cref{sec:cpinn_advantage}, their approach is similar to using the zero-sum game 
\begin{gather*}
    \min_{\bp} \max_{\bd} \log \left(\left(\bp^{\top} \bA \bd\right)^2\right) - \log\left(\left\|\bd\right\|^2\right).
\end{gather*}

\citet{wang2020understanding} and \citet{xu2021weak} adapted the penalty weights in PINNs during training to improve accuracy. 
Different from the $\cD$ output of CPINNs, these weights are multiplied with the \emph{square} violation of the equality constraint, which is always greater than zero.
Therefore, they do not correspond to a meaningful zero-sum game because the optimal discriminator strategy, in this case, drives all weights to infinity.
Alternatively, \citet{krishnapriyan2021characterizing} recommends the use of curriculum learning. 
They train the initial PINN on a better-conditioned variant of the original problem that slowly transforms into the target problem during training.

\section{Discussion}

This work introduced CPINNs, an agent-based approach to the neural-network-based solution of PDEs. 
CPINNs are crafted to avoid the ill-conditioning resulting from traditional PINNs least squares loss functions.
\Cref{sec:results} showed that CPINNs trained with ACGD improve upon the accuracy of PINNs trained with a comparable computational budget and can solve PDEs beyond even single-precision floating-point accuracy, the first approach with this capability.

With CPINNs, one can now achieve single-precision errors, but significant computation is still required.
This is due to the number of CG iterations performed for each ACGD step. 
Reducing this overhead is a direction for future work. 
Potential solutions to this problem include cheaper approximate solutions of the matrix inverse, different iterative solvers, or the recycling of Krylov subspaces~\citep{paige1975solution,parks2006recycling}.

The present experiments used the same training points for each iteration, following~\citet{RAISSI2019686}. 
In the future, we will investigate the effects of batch stochasticity on training accuracy.
We also plan to investigate competitive mirror descent (CMD) for partial differential inequalities and contact problems~\citep{schafer2020cmd,lions1972partial} and, more generally, CPINN-like approaches to other problems involving the constrained training of neural networks.

%
\subsubsection*{Acknowledgments}
This research was supported in part through research cyberinfrastructure resources and services provided by the Partnership for an Advanced Computing Environment (PACE) at the Georgia Institute of Technology, Atlanta, Georgia, USA. 
We thank Hongkai Zheng for maintaining the CGDs package used to conduct this work.

\bibliography{iclr2023_conference}

\begin{thebibliography}{54}
\providecommand{\natexlab}[1]{#1}
\providecommand{\url}[1]{\texttt{#1}}
\expandafter\ifx\csname urlstyle\endcsname\relax
  \providecommand{\doi}[1]{doi: #1}\else
  \providecommand{\doi}{doi: \begingroup \urlstyle{rm}\Url}\fi

\bibitem[Axelsson(1977)]{axelsson1977solution}
Owe Axelsson.
\newblock Solution of linear systems of equations: {I}terative methods.
\newblock In \emph{Sparse matrix techniques}, pp.\  1--51. Springer, 1977.

\bibitem[Bacon et~al.(2019)Bacon, Sch{\"a}fer, Gehring, Anandkumar, and
  Brunskill]{bacon2019lagrangian}
Pierre-Luc Bacon, Florian Sch{\"a}fer, Clement Gehring, Animashree Anandkumar,
  and Emma Brunskill.
\newblock A {L}agrangian method for inverse problems in reinforcement learning.
\newblock In \emph{Optimization in RL workshop at NeurIPS}, 2019.

\bibitem[Balduzzi et~al.(2018)Balduzzi, Racaniere, Martens, Foerster, Tuyls,
  and Graepel]{balduzzi2018mechanics}
David Balduzzi, Sebastien Racaniere, James Martens, Jakob Foerster, Karl Tuyls,
  and Thore Graepel.
\newblock The mechanics of n-player differentiable games.
\newblock In \emph{International Conference on Machine Learning}, pp.\
  354--363. PMLR, 2018.

\bibitem[Brezzi(1974)]{brezzi1974existence}
Franco Brezzi.
\newblock On the existence, uniqueness and approximation of saddle-point
  problems arising from {L}agrangian multipliers.
\newblock \emph{Publications math{\'e}matiques et informatique de Rennes},
  \penalty0 (S4):\penalty0 1--26, 1974.

\bibitem[Bubeck et~al.(2015)]{bubeck2015convex}
S{\'e}bastien Bubeck et~al.
\newblock Convex optimization: {A}lgorithms and complexity.
\newblock \emph{Foundations and Trends in Machine Learning}, 8\penalty0
  (3-4):\penalty0 231--357, 2015.

\bibitem[Daskalakis et~al.(2018)Daskalakis, Ilyas, Syrgkanis, and
  Zeng]{daskalakis2018training}
Constantinos Daskalakis, Andrew Ilyas, Vasilis Syrgkanis, and Haoyang Zeng.
\newblock Training {GAN}s with optimism.
\newblock In \emph{International Conference on Learning Representations}, 2018.

\bibitem[Daw et~al.(2022)Daw, Bu, Wang, Perdikaris, and
  Karpatne]{daw2022rethinking}
Arka Daw, Jie Bu, Sifan Wang, Paris Perdikaris, and Anuj Karpatne.
\newblock Rethinking the importance of sampling in physics-informed neural
  networks.
\newblock \emph{arXiv:2207.02338}, 2022.

\bibitem[Donti et~al.(2021)Donti, Rolnick, and Kolter]{donti2021dc}
Priya~L. Donti, David Rolnick, and J~Zico Kolter.
\newblock {DC}3: {A} learning method for optimization with hard constraints.
\newblock In \emph{International Conference on Learning Representations}, 2021.

\bibitem[E \& Yu(2017)E and Yu]{yu2017deep}
Weinan E and Bing Yu.
\newblock The {D}eep {R}itz method: {A} deep learning-based numerical algorithm
  for solving variational problems.
\newblock \emph{Communications in Mathematics and Statistics}, 6:\penalty0
  1--12, 2017.

\bibitem[Fioretto et~al.(2020)Fioretto, Hentenryck, Mak, Tran, Baldo, and
  Lombardi]{fioretto2020lagrangian}
Ferdinando Fioretto, Pascal~Van Hentenryck, Terrence~WK Mak, Cuong Tran,
  Federico Baldo, and Michele Lombardi.
\newblock {L}agrangian duality for constrained deep learning.
\newblock In \emph{Joint European Conference on Machine Learning and Knowledge
  Discovery in Databases}, pp.\  118--135. Springer, 2020.

\bibitem[Fortin \& Brezzi(1991)Fortin and Brezzi]{fortin1991mixed}
Michel Fortin and Franco Brezzi.
\newblock \emph{Mixed and hybrid finite element methods}, volume~3.
\newblock New York: Springer-Verlag, 1991.

\bibitem[Gemp \& Mahadevan(2018)Gemp and Mahadevan]{gemp2018global}
Ian Gemp and Sridhar Mahadevan.
\newblock Global convergence to the equilibrium of {GAN}s using variational
  inequalities.
\newblock \emph{arXiv:1808.01531}, 2018.

\bibitem[Gidel et~al.(2019)Gidel, Berard, Vincent, and
  Lacoste-Julien]{gidel2020variational}
Gauthier Gidel, Hugo Berard, Pascal Vincent, and Simon Lacoste-Julien.
\newblock A variational inequality perspective on generative adversarial nets.
\newblock \emph{arXiv:1802.10551}, 2019.

\bibitem[Kingma \& Ba(2014)Kingma and Ba]{kingma2017adam}
Diederik~P. Kingma and Jimmy Ba.
\newblock Adam: {A} method for stochastic optimization.
\newblock \emph{arXiv:1412.6980}, 2014.

\bibitem[Korpelevich(1977)]{korpelevich1977extragradient}
G.~M. Korpelevich.
\newblock Extragradient method for finding saddle points and other problems.
\newblock \emph{Matekon}, 13\penalty0 (4):\penalty0 35--49, 1977.

\bibitem[Krishnapriyan et~al.(2021)Krishnapriyan, Gholami, Zhe, Kirby, and
  Mahoney]{krishnapriyan2021characterizing}
Aditi~S. Krishnapriyan, Amir Gholami, Shandian Zhe, Robert~M. Kirby, and
  Michael~W. Mahoney.
\newblock Characterizing possible failure modes in physics-informed neural
  networks.
\newblock \emph{arXiv:2109.01050}, 2021.

\bibitem[Lagaris et~al.(1998)Lagaris, Likas, and
  Fotiadis]{lagaris1998artificial}
Isaac~E Lagaris, Aristidis Likas, and Dimitrios~I Fotiadis.
\newblock Artificial neural networks for solving ordinary and partial
  differential equations.
\newblock \emph{IEEE transactions on neural networks}, 9\penalty0 (5):\penalty0
  987--1000, 1998.

\bibitem[Lavin et~al.(2021)Lavin, Zenil, Paige, Krakauer, Gottschlich, Mattson,
  Anandkumar, Choudry, Rocki, Baydin, et~al.]{lavin2021simulation}
Alexander Lavin, Hector Zenil, Brooks Paige, David Krakauer, Justin
  Gottschlich, Tim Mattson, Anima Anandkumar, Sanjay Choudry, Kamil Rocki,
  At{\i}l{\i}m~G{\"u}ne{\c{s}} Baydin, et~al.
\newblock Simulation intelligence: {T}owards a new generation of scientific
  methods.
\newblock \emph{arXiv:2112.03235}, 2021.

\bibitem[Lemaire(1973)]{lemaire1973saddle}
B.~Lemaire.
\newblock Saddle-point problems in partial differential equations and
  applications to linear quadratic differential games.
\newblock \emph{Annali della Scuola Normale Superiore di Pisa-Classe di
  Scienze}, 27\penalty0 (1):\penalty0 105--160, 1973.

\bibitem[Letcher et~al.(2019)Letcher, Balduzzi, Racaniere, Martens, Foerster,
  Tuyls, and Graepel]{letcher2019differentiable}
Alistair Letcher, David Balduzzi, S{\'e}bastien Racaniere, James Martens, Jakob
  Foerster, Karl Tuyls, and Thore Graepel.
\newblock Differentiable game mechanics.
\newblock \emph{The Journal of Machine Learning Research}, 20\penalty0
  (1):\penalty0 3032--3071, 2019.

\bibitem[Li et~al.(2021)Li, Zheng, Kovachki, Jin, Chen, Liu, Azizzadenesheli,
  and Anandkumar]{li2021physics}
Zongyi Li, Hongkai Zheng, Nikola Kovachki, David Jin, Haoxuan Chen, Burigede
  Liu, Kamyar Azizzadenesheli, and Anima Anandkumar.
\newblock Physics-informed neural operator for learning partial differential
  equations.
\newblock \emph{arXiv:2111.03794}, 2021.

\bibitem[Liao \& Ming(2019)Liao and Ming]{liao2019deep}
Yulei Liao and Pingbing Ming.
\newblock Deep {N}itsche method: {D}eep {R}itz method with essential boundary
  conditions.
\newblock \emph{arXiv:1912.01309}, 2019.

\bibitem[Lions(1972)]{lions1972partial}
JL~Lions.
\newblock Partial differential inequalities.
\newblock \emph{Russian Mathematical Surveys}, 27\penalty0 (2):\penalty0 91,
  1972.

\bibitem[Liu et~al.(2021)Liu, Chen, Du, and Tegmark]{liu2021physicsaugmented}
Ziming Liu, Yunyue Chen, Yuanqi Du, and Max Tegmark.
\newblock {P}hysics-{A}ugmented {L}earning: A new paradigm beyond
  {P}hysics-{I}nformed {L}earning.
\newblock \emph{arXiv:2109.13901}, 2021.

\bibitem[Lokhande et~al.(2020)Lokhande, Akash, Ravi, and
  Singh]{lokhande2020fairalm}
Vishnu~Suresh Lokhande, Aditya~Kumar Akash, Sathya~N Ravi, and Vikas Singh.
\newblock Fair{ALM}: Augmented {L}agrangian method for training fair models
  with little regret.
\newblock In \emph{European Conference on Computer Vision}, pp.\  365--381.
  Springer, 2020.

\bibitem[Lu et~al.(2021)Lu, Meng, Mao, and Karniadakis]{lu2021deepxde}
Lu~Lu, Xuhui Meng, Zhiping Mao, and George~Em Karniadakis.
\newblock Deep{XDE}: {A} deep learning library for solving differential
  equations.
\newblock \emph{SIAM Review}, 63\penalty0 (1):\penalty0 208--228, 2021.

\bibitem[Mescheder et~al.(2017)Mescheder, Nowozin, and
  Geiger]{mescheder2017numerics}
Lars Mescheder, Sebastian Nowozin, and Andreas Geiger.
\newblock The numerics of {GANs}.
\newblock \emph{Advances in neural information processing systems}, 30, 2017.

\bibitem[Nabian et~al.(2021)Nabian, Gladstone, and
  Meidani]{nabian2021efficient}
Mohammad~Amin Nabian, Rini~Jasmine Gladstone, and Hadi Meidani.
\newblock Efficient training of physics-informed neural networks via importance
  sampling.
\newblock \emph{Computer-Aided Civil and Infrastructure Engineering},
  36\penalty0 (8):\penalty0 962--977, 2021.

\bibitem[Owhadi(2017)]{owhadi2017multigrid}
Houman Owhadi.
\newblock Multigrid with rough coefficients and multiresolution operator
  decomposition from hierarchical information games.
\newblock \emph{SIAM Review}, 59\penalty0 (1):\penalty0 99--149, 2017.

\bibitem[Paige \& Saunders(1975)Paige and Saunders]{paige1975solution}
Christopher~C Paige and Michael~A Saunders.
\newblock Solution of sparse indefinite systems of linear equations.
\newblock \emph{SIAM Journal on Numerical Analysis}, 12\penalty0 (4):\penalty0
  617--629, 1975.

\bibitem[Parks et~al.(2006)Parks, De~Sturler, Mackey, Johnson, and
  Maiti]{parks2006recycling}
Michael~L Parks, Eric De~Sturler, Greg Mackey, Duane~D Johnson, and Spandan
  Maiti.
\newblock Recycling {K}rylov subspaces for sequences of linear systems.
\newblock \emph{SIAM Journal on Scientific Computing}, 28\penalty0
  (5):\penalty0 1651--1674, 2006.

\bibitem[Pathak et~al.(2015)Pathak, Krahenbuhl, and
  Darrell]{pathak2015constrained}
Deepak Pathak, Philipp Krahenbuhl, and Trevor Darrell.
\newblock Constrained convolutional neural networks for weakly supervised
  segmentation.
\newblock In \emph{Proceedings of the IEEE International Conference on Computer
  Vision}, pp.\  1796--1804, 2015.

\bibitem[Quarteroni \& Valli(2008)Quarteroni and
  Valli]{quarteroni2008numerical}
Alfio Quarteroni and Alberto Valli.
\newblock \emph{Numerical approximation of partial differential equations},
  volume~23.
\newblock Springer Science \& Business Media, 2008.

\bibitem[Raissi et~al.(2019)Raissi, Perdikaris, and Karniadakis]{RAISSI2019686}
Maziar Raissi, Paris Perdikaris, and George~Em Karniadakis.
\newblock Physics-informed neural networks: {A} deep learning framework for
  solving forward and inverse problems involving nonlinear partial differential
  equations.
\newblock \emph{Journal of Computational Physics}, 378:\penalty0 686--707,
  2019.
\newblock ISSN 0021-9991.

\bibitem[Ruder(2016)]{ruder2016overview}
Sebastian Ruder.
\newblock An overview of gradient descent optimization algorithms.
\newblock \emph{arXiv:1609.04747}, 2016.

\bibitem[Saad \& Schultz(1986)Saad and Schultz]{saad1986gmres}
Youcef Saad and Martin~H Schultz.
\newblock {GMRES}: {A} generalized minimal residual algorithm for solving
  nonsymmetric linear systems.
\newblock \emph{SIAM Journal on Scientific and Statistical Computing},
  7\penalty0 (3):\penalty0 856--869, 1986.

\bibitem[Sch{\"a}fer \& Anandkumar(2019)Sch{\"a}fer and
  Anandkumar]{schafer2020competitive}
Florian Sch{\"a}fer and Anima Anandkumar.
\newblock Competitive gradient descent.
\newblock In \emph{NeurIPS}, 2019.

\bibitem[Sch{\"a}fer et~al.(2020{\natexlab{a}})Sch{\"a}fer, Anandkumar, and
  Owhadi]{schafer2020cmd}
Florian Sch{\"a}fer, Anima Anandkumar, and Houman Owhadi.
\newblock Competitive {M}irror {D}escent.
\newblock \emph{arXiv:2006.10179}, 2020{\natexlab{a}}.

\bibitem[Sch{\"a}fer et~al.(2020{\natexlab{b}})Sch{\"a}fer, Zheng, and
  Anandkumar]{schafer2020implicit}
Florian Sch{\"a}fer, Hongkai Zheng, and Anima Anandkumar.
\newblock Implicit competitive regularization in {GANs}.
\newblock \emph{arXiv:1910.05852}, 2020{\natexlab{b}}.

\bibitem[Shang et~al.(2022)Shang, Wang, and Sun]{shang2022deep}
Yong Shang, Fei Wang, and Jingbo Sun.
\newblock Deep {P}etrov-{G}alerkin method for solving partial differential
  equations.
\newblock \emph{arXiv:2201.12995}, 2022.

\bibitem[Shewchuk(1994)]{shewchuk1994introduction}
Jonathan~Richard Shewchuk.
\newblock An introduction to the conjugate gradient method without the
  agonizing pain.
\newblock Technical report, Carnegie Mellon University, 1994.

\bibitem[Soliman et~al.(2021)Soliman, Chern, Diamanti, Kn{\"o}ppel, Pinkall,
  and Schr{\"o}der]{soliman2021constrained}
Yousuf Soliman, Albert Chern, Olga Diamanti, Felix Kn{\"o}ppel, Ulrich Pinkall,
  and Peter Schr{\"o}der.
\newblock Constrained {W}illmore surfaces.
\newblock \emph{ACM Transactions on Graphics (TOG)}, 40\penalty0 (4):\penalty0
  1--17, 2021.

\bibitem[van~der Meer et~al.(2022)van~der Meer, Oosterlee, and
  Borovykh]{van2022optimally}
Remco van~der Meer, Cornelis~W Oosterlee, and Anastasia Borovykh.
\newblock Optimally weighted loss functions for solving {PDEs} with neural
  networks.
\newblock \emph{Journal of Computational and Applied Mathematics},
  405:\penalty0 113887, 2022.

\bibitem[Wald(1945)]{wald1945statistical}
Abraham Wald.
\newblock Statistical decision functions which minimize the maximum risk.
\newblock \emph{Annals of Mathematics}, pp.\  265--280, 1945.

\bibitem[Wang et~al.(2021{\natexlab{a}})Wang, Teng, and
  Perdikaris]{wang2020understanding}
Sifan Wang, Yujun Teng, and Paris Perdikaris.
\newblock Understanding and mitigating gradient pathologies in physics-informed
  neural networks.
\newblock \emph{SIAM Journal of Scientific Computing}, 43:\penalty0
  A3055--A3081, 2021{\natexlab{a}}.

\bibitem[Wang et~al.(2021{\natexlab{b}})Wang, Wang, and
  Perdikaris]{wang2021learning}
Sifan Wang, Hanwen Wang, and Paris Perdikaris.
\newblock Learning the solution operator of parametric partial differential
  equations with physics-informed {DeepONets}.
\newblock \emph{Science advances}, 7\penalty0 (40), 2021{\natexlab{b}}.

\bibitem[Wang et~al.(2022{\natexlab{a}})Wang, Sankaran, and
  Perdikaris]{wang2022respecting}
Sifan Wang, Shyam Sankaran, and Paris Perdikaris.
\newblock Respecting causality is all you need for training physics-informed
  neural networks.
\newblock \emph{arXiv:2203.07404}, 2022{\natexlab{a}}.

\bibitem[Wang et~al.(2022{\natexlab{b}})Wang, Yu, and Perdikaris]{wang2022and}
Sifan Wang, Xinling Yu, and Paris Perdikaris.
\newblock When and why {PINNs} fail to train: {A} neural tangent kernel
  perspective.
\newblock \emph{Journal of Compuational Physics}, 449:\penalty0 110768,
  2022{\natexlab{b}}.

\bibitem[Wight \& Zhao(2020)Wight and Zhao]{wight2020solving}
Colby~L Wight and Jia Zhao.
\newblock Solving {A}llen--{C}ahn and {C}ahn--{H}illiard equations using the
  adaptive physics informed neural networks.
\newblock \emph{arXiv:2007.04542}, 2020.

\bibitem[Xu et~al.(2021)Xu, Zhang, Rong, and Wang]{xu2021weak}
Rui Xu, Dongxiao Zhang, Miao Rong, and Nanzhe Wang.
\newblock Weak form theory-guided neural network ({TgNN-wf}) for deep learning
  of subsurface single-and two-phase flow.
\newblock \emph{Journal of Computational Physics}, 436:\penalty0 110318, 2021.

\bibitem[Xue et~al.(2020)Xue, Beatson, Adriaenssens, and
  Adams]{xue2020amortized}
Tianju Xue, Alex Beatson, Sigrid Adriaenssens, and Ryan Adams.
\newblock Amortized finite element analysis for fast {PDE}-constrained
  optimization.
\newblock In \emph{International Conference on Machine Learning}, pp.\
  10638--10647. PMLR, 2020.

\bibitem[Yu et~al.(2021)Yu, Gehring, Sch{\"a}fer, and Anandkumar]{yu2021robust}
Jing Yu, Clement Gehring, Florian Sch{\"a}fer, and Animashree Anandkumar.
\newblock Robust reinforcement learning: {A} constrained game-theoretic
  approach.
\newblock In \emph{Learning for Dynamics and Control}, pp.\  1242--1254, 2021.

\bibitem[Zang et~al.(2020)Zang, Bao, Ye, and Zhou]{zang2020weak}
Yaohua Zang, Gang Bao, Xiaojing Ye, and Haomin Zhou.
\newblock Weak adversarial networks for high-dimensional partial differential
  equations.
\newblock \emph{Journal of Computational Physics}, 411:\penalty0 109409, 2020.

\bibitem[Zheng(2020)]{cgds-package}
Hongkai Zheng.
\newblock {CGD}s.
\newblock \url{https://github.com/devzhk/cgds-package}, 2020.

\end{thebibliography}
\bibliographystyle{iclr2023_conference}

\appendix
\section{Appendix}

This appendix contains details on the experiments mentioned in \Cref{sec:results}.

All PINN models and CPINN generators use the hyperbolic tangent as the activation function. Discriminators in CPINNs contain ReLU as the activation function.
The number of neurons and layers in the neural networks varies for each equation.
We use the GMRES-based ACGD implementation of \citet{cgds-package}, available under the MIT license at \url{https://github.com/devzhk/cgds-package}, as well as the implementation of ExtraGradient methods available under the MIT license at \url{https://github.com/GauthierGidel/Variational-Inequality-GAN}.
For the experiments in \cref{s:BurgersSection}, \ref{s:ACsection},\ref{s:SchrodingerSection} we use the training and testing data sets from \url{https://github.com/maziarraissi/PINNs/tree/master/main}, which are available under the MIT license.
 
\subsection{Poisson Equation}

For the Poisson equation, we use $5\ 000$ training points within the domain $[-2, 2]\times[-2,2]$, $50$ training points on each side of the domain boundary. 
We randomly selected all training points with Latin Hypercube sampling. 
The PINN model contains 3 hidden layers with $50$ neurons in each layer. 
The discriminator in the CPINN contains 4 hidden layers with 50 neurons in each layer.

The PINNs are trained with Adam~\citep{kingma2017adam} and SGD~\citep{ruder2016overview}.
Adam and ACGD both use a learning rate of $10^{-3}$, beta values $\beta_1 = 0.99$ and $\beta_2 = 0.99$. 
The $\epsilon$ of Adam and ACGD are each set to $10^{-8}$ and $10^{-6}$, resepctively (all parameters following the usual naming conventions).


\subsection{Schr\"odinger's Equation}

We use Latin Hypercube sampling to randomly select $20\ 000$ training points within the domain and $50$ points on each boundary.

We test several different network configurations and hyperparameters. 
This includes combinations of number of neurons per layer for each network, including 100 and 200 neurons per layer; learning rates of $10^{-5}, 10^{-4}, 5\times 10^{-4}, 10^{-3}, 10^{-2}$, and $2\times 10^{-2}$;  and Adam and ACGD $\beta$ values of $(0.99,0.99)$ and $(0.9, 0.999)$.

We chose $10^{-4}$ and $10^{-3}$ as the learning rate, $(0.9, 0.999)$ and $(0.99, 0.99)$ as the $\beta$ values, $10^{-8}$ and $10^{-6}$ as the $\epsilon$ for Adam and ACGD, respectively.
The iterative linear solve of the ACGD optimizer uses a relative tolerance of $10^{-7}$ and absolute tolerance of $10^{-20}$.

The PINN presented in \cref{fig:SchrodingerPlot} contains 4 hidden layers with 100 neurons per layer, and the discriminator in CPINN contains 4 hidden layers with 200 neurons per layer.




\subsection{Burgers' Equation}

For the Burgers' equation, we use Latin Hypercube sampling to randomly select $10\ 000$ training points within the domain, $100$ points at the initial condition and $100$ points for each boundary condition specified in \cref{BurgersBC}.

The PINN model contains 8 hidden layers with 60 neurons per layer, and the discriminator in the CPINN contains 8 hidden layers with 60 neurons per layer.

The $\beta$ of Adam and ACGD are $(0.99, 0.99)$, and the learning rates of both optimizers are $10^{-3}$.
The $\epsilon$ are $10^{-8}$ and $10^{-6}$ for Adam and ACGD, respectively.
The iterative linear solve of the ACGD optimizer had a relative tolerance of $10^{-7}$ and an absolute tolerance of $10^{-20}$.

\subsection{Allen-Cahn Equation}

For the Allen Cahn equation, we randomly select $10\ 000$ training points within the domain for the PDE constraint in \cref{e:ac}, $100$ and $256$ points for the initial condition and boundary condition specified in \cref{e:acbc}, respectively.
The learning rates of Adam and ACGD are both $10^{-3}$, the $\beta$ of Adam and ACGD are $(0.99, 0.99)$, and the $\epsilon$ are $10^{-8}$, $10^{-6}$, respectively.

The PINN model contains 4 hidden layers with 128 neurons per layer for the adaptive sampling method experiments. 
The discriminator in the CPINN contains 4 hidden layers with 256 neurons per layer. 
GMRES was used as the inner iterative solver in ACGD, with a relative tolerance of $10^{-7}$ and an absolute tolerance of $10^{-20}$.

For the curriculum learning approach, we divide the $10\ 000$ points into $10$ subsets based on the time coordinate. 
Once the $\PINNPDELoss$ on the current training set is less than $10^{-7}$, we include the next subset of collocation points in the training.

\subsection{Other optimizers}
In addition to ACGD and Adam, we also use SGD, CGD~\citep{schafer2020competitive}, 
ExtraAdam and ExtraSGD~\citep{gidel2020variational} to approximate the solution of~\eqref{e:minimax} in \cref{s:optimizers}. 
We test several hyperparameter combinations for each optimizer. 
The learning rates vary included $10^{-4}, 5\times 10^{-4}, 10^{-3}, 10^{-2}$, and $2\times 10^{-2}$. 
The $\beta$ of Adam and ACGD are $(0.99,0.99)$. 
We tested several $\beta$ values for Adam and ACGD but did not observe meaningful changes to our results or conclusions when varying them.
For the ExtraAdam optimizer, we tested several pairs of $\beta$ values, from $0.3, 0.5, 0.7, 0.9, 0.99$, though none of them provide accuracy competitive with the other optimizers.
For CGD, SGD and ExtraSGD, the learning rates are $10^{-2}$.
The CGD optimizer uses conjugate gradient as the iterative solver with relative tolerance $10^{-12}$ and absolute tolerance $10^{-20}$.
\end{document}